\documentclass[11pt]{article}

\usepackage[final]{acl}

\usepackage{times}
\usepackage{latexsym}

\usepackage[T1]{fontenc}

\usepackage[utf8]{inputenc}

\usepackage{microtype}

\usepackage{inconsolata}

\usepackage{graphicx}
\usepackage{multirow}
\usepackage{calc}
 \usepackage{booktabs}
\usepackage{longtable}

\usepackage{listings}
\usepackage{textcomp} 
\usepackage{xcolor}

\title{ClinStructor: AI-Powered Structuring of Unstructured Clinical Texts}

\author{Karthikeyan K \\ Duke University \\ karthikeyan.k@duke.edu
        \AND
        Raghuveer Thirukovalluru \\ Duke University \\ raghuveer.thirukovalluru@duke.edu
        \And
        David Carlson\\ Duke University \\ david.carlson@duke.edu }

\begin{document}

\maketitle

\begin{abstract}
Clinical notes contain valuable, context-rich information, but their unstructured format introduces several challenges, including unintended biases (e.g., gender or racial bias), and poor generalization across clinical settings (e.g., models trained on one EHR system may perform poorly on another due to format differences) and poor interpretability. To address these issues, we present ClinStructor, a pipeline that leverages large language models (LLMs) to convert clinical free-text into structured, task-specific question–answer pairs prior to predictive modeling. Our method substantially enhances transparency and controllability and only leads to a modest reduction in predictive performance (a 2–3\% drop in AUC), compared to direct fine-tuning, on the ICU mortality prediction task. ClinStructor lays a strong foundation for building reliable, interpretable, and generalizable machine learning models in clinical environments.
\end{abstract}

\section{Introduction}

Recent advances in Artificial Intelligence (AI) have driven significant progress across various domains, including healthcare~\citep{cascella2023evaluating,yang2024talk2care}, finance~\citep{yu2023temporal,li2023large,zhao2024revolutionizing}, law~\citep{siino2025exploring}. However, in high-stakes contexts such as clinical decision-making, requirements extend beyond prediction performance alone. It becomes critical not only to develop reliable predictive models but also to maintain fine-grained control over the attributes influencing predictions. In such scenarios, interpretability is not merely desirable—it is often essential to comply with regulations (e.g., not using protected features such as gender, occupation, etc.) and to foster trust among clinicians, patients, and stakeholders.


Clinical notes, a ubiquitous component of electronic health record (EHR) systems, constitute one of the richest sources of clinical information. These free-text notes can encapsulate nuanced details, thus providing invaluable data for predictive modeling. Nonetheless, clinical free-text is inherently unstructured, presenting both significant opportunities and substantial challenges. Notably, clinical notes lack a standardized format, varying extensively in style, structure, and terminology—not only internationally and across languages but even within individual hospitals between different departments or clinicians~\citep{cohen2019variation}. This variability complicates the development of reliable and controllable predictive models.

Training predictive models directly on clinical text in a black-box manner raises important concerns. Firstly, there is often insufficient insight into the exact information leveraged by the model to predict an outcome. Clinical notes frequently contain protected attributes, such as race or gender, which -- while potentially relevant in certain contexts -- may inadvertently introduce biases if not explicitly managed. Additionally, clinical text is susceptible to label leakage, especially for tasks such as mortality prediction, where phrases like “the patient has passed away” or “in critical condition” can directly reveal outcomes. Consequently, models might learn superficial shortcuts, inflating performance metrics without capturing meaningful clinical signals. Moreover, models optimized on a test set from the same data distribution as training may fail to generalize effectively to external datasets from differing institutions, regions, or populations. Variations in documentation practices, terminology, and structure can potentially degrade performance in real-world, cross-site deployments.

\begin{figure*} 
    \centering 
    \includegraphics[width=.95\linewidth]{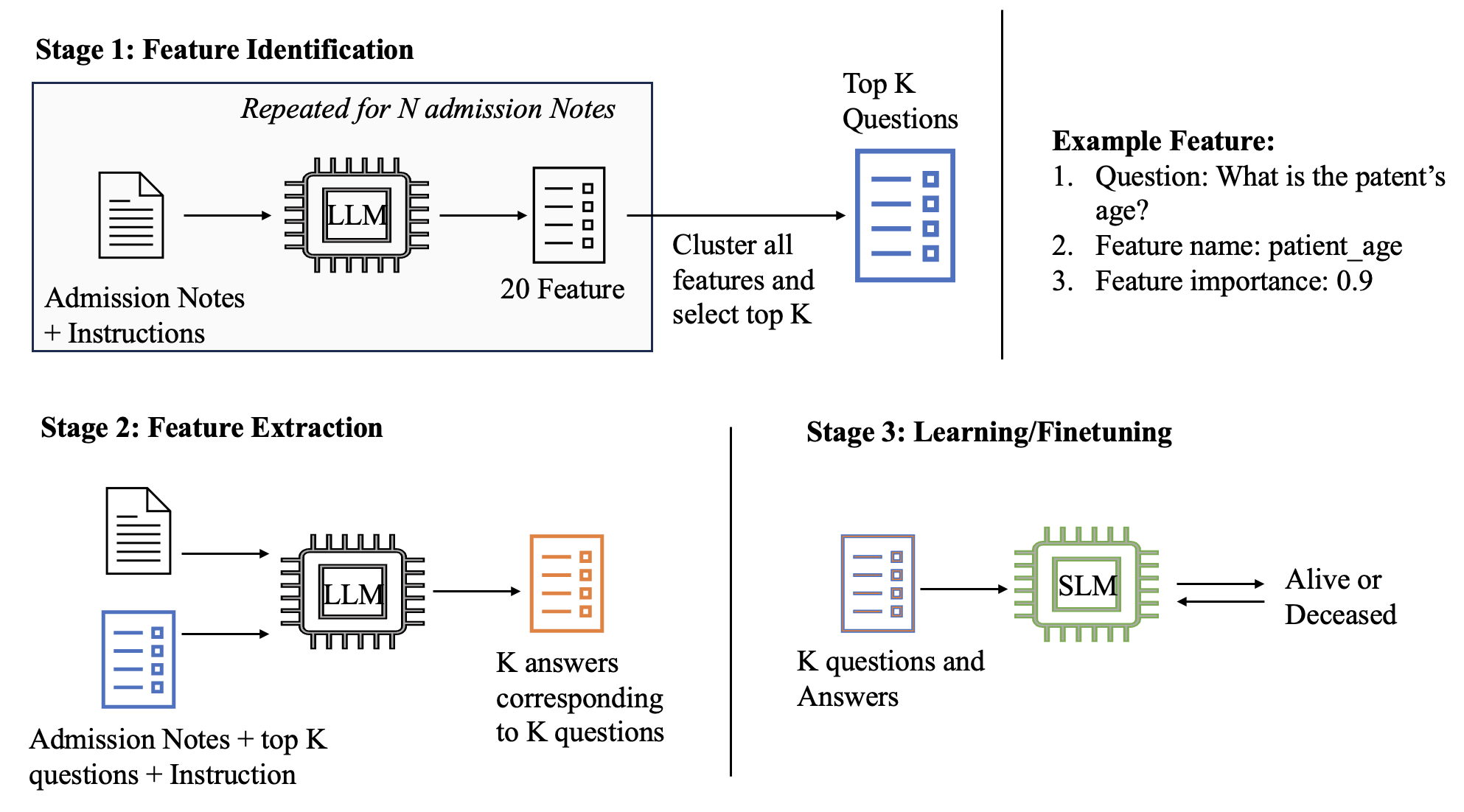} 
    \caption{\textbf{Illustration of the ClinStructor Pipeline:} The proposed pipeline comprises three main stages: (1) Feature Identification, (2) Feature Extraction, and (3) fine-tuning.} \label{fig:pipeline} 
\end{figure*}

Several approaches have been proposed to mitigate the challenges posed by unstructured clinical text. These include: (1) bias mitigation methods, which either remove sensitive attributes such as race or gender from input data or transform input to an intermediate representations that obscure sensitive information~\citep{sun2019mitigatinggenderbiasnatural}; (2) distributional robustness techniques, explicitly designed to improve resilience against shifts in data distributions across EHR databases or clinical institutes or patient populations~\citep{tan-etal-2022-domain}; and (3) post-hoc explanation techniques, such as highlighting textual spans most influential to predictions~\citep{vig-2019-multiscale}. While valuable, these methods frequently fail to address the root cause of many challenges -- the fundamentally unstructured nature of clinical text, making it inherently difficult to ascertain which information the model has utilized.


To address the underlying issue of clinical text’s inherent lack of structure, we introduce \textbf{ClinStructor}, illustrated in Figure~\ref{fig:pipeline}. ClinStructor is a novel pipeline designed to transform clinical notes into structured representations (question-answer pairs), prior to predictive modeling. Instead of directly fine-tuning models on raw notes, we leverage LLMs to systematically extract meaningful, task-relevant features, in our case, ICU mortality prediction. Specifically, we prompt LLMs to identify a set of clinically relevant questions informed by both clinical domain knowledge and by task-specific data. Subsequently, the same LLMs extract answers to these questions from the patient's admission notes, resulting in structured question-answer pairs. This effectively converts the unstructured clinical text into structured data, significantly enhancing transparency and controllability. By standardizing the input format, our approach enables thorough inspection of the extracted features, facilitating verification of their clinical relevance and consistency. While subsequent fine-tuning of predictive models on these structured representations slightly diminishes interpretability, it allows leveraging pretrained LLM knowledge for the downstream predictive task. Thus, our approach bridges the expressive richness of unstructured text with the interpretability, reliability, and generalizability inherent in structured data representations.

We empirically evaluate ClinStructor against direct fine-tuning on raw clinical notes. Our results indicate a modest performance trade-off, a decrease of approximately 2 to 3\% in AUC. The observed performance reduction is anticipated due to potential information loss inherent to structured transformations. Nevertheless, this trade-off delivers substantial advantages. Explicitly defining features used for prediction enables inspection and verification of input signals, effectively mitigating label leakage concerns. Importantly, our approach provides foundational building blocks for constructing predictive models exhibiting enhanced interpretability, robustness, and generalizability—especially crucial when deployed across diverse clinical environments characterized by substantial variability in documentation practices.

The remainder of this paper is structured as follows. Section~\ref{sec:related_work} discusses background and related literature. Section~\ref{sec:method} details our ClinStructor pipeline, describing each stage comprehensively. Section~\ref{sec:experiments} outlines our experimental setup and presents comparative quantitative results. Further analytical insights are provided in Section~\ref{sec:analysis}. In Section~\ref{sec:discussion}, we highlight our method’s limitations and propose mitigation strategies. Finally, Section~\ref{sec:conclusion} summarizes key insights and concludes.

The primary contributions of our work include:
\begin{enumerate} 
\item Introducing \textbf{ClinStructor}, a novel LLM-based pipeline converting unstructured clinical notes into structured, task-specific question-answer pairs. 
\item Empirically demonstrating that ClinStructor retains the majority of clinically relevant information, achieving competitive predictive performance in the ICU Mortality prediction task. 
\item Discussing crucial limitations of our structured approach, along with detailed suggestions for addressing these issues in future research. 
\end{enumerate}



\section{Background and Related Works}\label{sec:related_work}

\subsection{Overview of Interpretability Techniques}

Interpretability is a crucial and growing focus within Machine Learning research~\citep{molnar2020interpretable,du2019techniques}. Traditional models, such as linear regression, Support Vector Machines, decision trees, and AdaBoost, inherently provide interpretability for tabular data (numerical and categorical). However, interpretability significantly diminishes with more complex modalities, such as textual or image data, which typically require more sophisticated modeling approaches.

For text-based data, several post-hoc methods have been proposed to explain model predictions. Notable examples include SHapley Additive exPlanations (SHAP)\citep{NIPS2017_8a20a862}, LIME for Text\citep{ribeiro2016whyitrustyou}, Saliency Maps~\citep{simonyan2014deepinsideconvolutionalnetworks}, attention visualization techniques~\citep{vig-2019-multiscale,tenney2020language}, and Integrated Gradients~\citep{sundararajan2017axiomaticattributiondeepnetworks}. These approaches primarily rely on highlighting influential segments of text to explain predictions. However, these explanations may not be faithful or reliable.

Alternatively, interpretability can also be achieved through similarity or example-based methods. Techniques such as K-Nearest Neighbors (K-NN) applied to text embeddings, Prototypical Networks~\citep{snell2017prototypicalnetworksfewshotlearning}, Case-Based Reasoning~\citep{wiratunga2024cbrragcasebasedreasoningretrieval}, counterfactual explanations~\citep{wachter2018counterfactualexplanationsopeningblack}, TracIn~\citep{pruthi2020estimatingtrainingdatainfluence,k2021revisitingmethodsfindinginfluential}, and Influence Functions~\citep{koh2020understandingblackboxpredictionsinfluence} provide interpretability by relating predictions to influential or similar training examples. These explanation methods are often very unstable~\citep{k2021revisitingmethodsfindinginfluential}.

\subsection{A Broad Overview of Applications of LLMs in Healthcare}

In recent years, there has been significant interest in applying Large Language Models (LLMs) to healthcare. Early studies~\citep{thirunavukarasu2023large,yang2023large,nazi2024large,nassiri2024recent} investigated the potential of LLMs in clinical practice, highlighting both opportunities and key challenges. Further research~\citep{denecke2024potential} has collected clinician perspectives, providing insights into practical strengths and limitations when deploying these models.

LLMs have demonstrated effectiveness across various clinical tasks. For example, they have been used to detect and anonymize Personally Identifiable Information (PII) within clinical notes~\citep{liu2024evaluating,k-etal-2023-taxonomy}. They can also effectively perform information extraction tasks from unstructured text~\citep{agrawal-etal-2022-large,lopez2025clinical}.
Moreover, LLMs have been evaluated for diagnostic support~\citep{panagoulias2023evaluating,panagoulias2024evaluating}, summarizing medical records~\citep{madzime2024enhanced,goodman2024ai}, clinical question answering~\citep{wang2023augmenting,li2024mediq}, medical coding~\citep{soroush2024large,li2024exploring}, and predictive modeling of patient outcomes~\citep{lyu2023multimodaltransformerfusingclinical,wu2023fineehrrefineclinicalnote,vanAken2021,rohr-etal-2024-revisiting}. Beyond clinical tasks, LLMs have also shown promise in medical education and training~\citep{safranek2023role,abd2023large,lucas2024systematic}.

\subsection{Specialized LLMs in Healthcare and Medical domain}

To better address healthcare specific applications, several domain-adapted LLMs have been proposed. Notable examples include Med-BERT~\citep{rasmy2021med}, ClinicalBERT~\citep{clinicalbert}, and BioBERT~\citep{lee2020biobert}, all pretrained on specialized clinical and biomedical datasets. More recent models like Meditron~\citep{chen2023meditron,bosselut2024meditron}, Med-PaLM~\citep{singhal2022largelanguagemodelsencode}, and BioMistral~\citep{labrak2024biomistralcollectionopensourcepretrained} further aim to improve performance on medical and clinical benchmarks. In section~\ref{sec:experiments}, we use Meditron-based models (Meditron Qwen2.5 7B and Meditron LLaMA 3.1 8B) for our experiments, as these are comparable to their general-purpose counterparts. 

\subsection{LLMs for Feature Importance}
Recent works have experimented with applying LLM-based approach for feature engineering, primarily targeting tabular datasets~\citep{tsymbalov2024llmfeatures,zhang2024dynamicadaptivefeaturegeneration}. In addition, other works have explored leveraging LLMs for feature selection and importance estimation, particularly within zero-shot settings~\citep{Jeong2024LLMSelectFS,choi2022lmpriorspretrainedlanguagemodels}. Similar to these studies, our approach utilizes LLMs not only to select but also to generate meaningful features, assigning them appropriate importance scores. 
 
\subsection{Text Bottleneck Model}

Relatively few inherently interpretable methods exist for textual data. One notable example is the Text Bottleneck Model (TBM) proposed by \citet{ludan2024interpretablebydesigntextunderstandingiteratively}. Inspired by Concept Bottleneck Models \citep{koh2020conceptbottleneckmodels}, TBM employs large language models (LLMs) to iteratively extract key textual concepts (e.g., “service” or “food quality” in sentiment classification) and builds interpretable models based on these concepts. TBM's methodology is iterative: in each round, the model is trained on the full dataset, and high-loss examples are used to extract additional concepts. This process is repeated multiple times, making it computationally expensive. Due to this, TBM faces significant scalability challenges. Its evaluation was limited to simple tasks such as sentiment analysis, using only small datasets (approximately 250 examples).  In contrast, our proposed approach extracts all relevant features in a single step, offering a scalable and structured alternative to TBM. The scalability of our proposed approach makes it suitable for more complex, real-world tasks.

\begin{table*}
  \centering
  \begin{tabular}{p{1cm}p{11cm}}
    \toprule
     Rank & Question  \\
    \midrule
    \multicolumn{2}{c}{Qwen 32b Instruct} \\
    \midrule
    1 & what is the patient’s age? \\
    2 & what medications was the patient on at the time of admission? \\
    3 & what is the patient's vital signs on admission?\\
    4 & what is the patient’s physical exam findings? \\
    5 & what are the patient's significant past medical conditions? \\
    \midrule
    \multicolumn{2}{c}{LLaMA 70B Instruct} \\
    \midrule 
     1 & what is the patient's medical history? \\
    2 & what is the patient's blood pressure on admission?\\
    3 & what is the patient's current level of cognitive function?\\
    4 & what are the patient's current medications and dosages?\\
    5 & what is the patient's age? \\
    \bottomrule
  \end{tabular}
  \caption{Top 5 questions generated by Qwen 32B and LLaMA 70B Instruct models}
  \label{tab:top_k_questions}
\end{table*}

\section{Our Proposed Method: ClinStructor}\label{sec:method} 
In Figure~\ref{fig:pipeline}, we illustrate our proposed method, ClinStructor, which converts unstructured clinical notes into structured representations, thereby enhancing their interpretability and control over the attributes used for downstream tasks. Our method involves three key steps: (1) Feature Identification, (2) Feature Extraction, and (3) Fine-tuning. Below, we detail each step thoroughly.

\subsection{Stage 1: Feature Identification} 
The first step, Feature Identification, establishes the foundation for the structured representation of clinical notes. Our goal is to discover a meaningful set of features that can be extracted from free-text admission notes and utilized for predictive modeling; specifically, we focus on ICU mortality prediction in this study. This step leverages both the large language model’s (LLM) inherent world knowledge as well as the information in examples from downstream predictive task. 

We begin by randomly sampling 1,000 ICU admission notes, equally balanced between positive (mortality) and negative (survival) outcomes. For each note, we prompt an LLM to generate a set of 20 potential candidate features. Each feature consists of three attributes:

\begin{itemize} \item 
    \textbf{Question}: A natural language question whose answer serves as the feature value. We instruct the LLM to produce generalizable questions (e.g., "What is the patient's age?" rather than "Is the patient's age 60?"). 
    \item \textbf{Feature Name}: A concise, one-word descriptor for the feature (e.g., age, medication, etc.). This primarily aids in subsequent clustering and de-duplication. 
    \item \textbf{Importance Score}: A numeric value between 0 and 1, reflecting the LLM's estimated importance of the feature for the downstream predictive task. 
\end{itemize}

This process yields a total of 20,000 candidate features (20 features per note × 1,000 notes). We have to select the top K(= 50) features. However, due to natural variation in language, many generated questions and feature names differ slightly despite having similar semantics. For example, "What is the patient's current age?" vs. "What is the patient's age?", or feature names like "current\_age" vs. "age". Consequently, simple strategies such as frequency-based selection would incorrectly segment semantically equivalent features, thereby underestimating their true frequency/importance.

To address this issue, we perform single-linkage clustering~\citep{8862232} over the set of generated features. Each node in the clustering graph represents one LLM-generated feature candidate (i.e., a question, a feature name, and an importance score). We define an edge between two nodes if they share either the exact same question or the exact same feature name (after basic normalization steps such as lower casing). Single-linkage clustering on this graph results in clusters of semantically equivalent questions.

Next, we compute the cluster weight as the sum of LLM-generated feature importance scores for all nodes within a cluster. We then select the top K clusters (K = 50) based on these aggregate cluster weights to form our final set of features. Manual inspection revealed no semantically duplicate questions, indicating that the clustering process effectively ensures question uniqueness. Within each selected cluster, we identify a representative question—specifically, the one with the highest cumulative importance score, calculated as the sum of feature importance scores across all exact occurrences of that question (not the feature names). This process results in a curated list of 50 unique, LLM-generated questions, ranked by their aggregate importance scores. These questions define the structured format used for subsequent feature extraction. It is important to note that we utilize downstream task data in two ways: (1) as input to the LLM to generate candidate features, and (2) during clustering, to select the top K features based on importance. Meanwhile, the LLM's internal knowledge contributes to generating relevant questions, feature names, and corresponding importance scores. Table~\ref{tab:top_k_questions} presents the top 5 generated questions from each of the Qwen2.5 32B Instruct and LLaMA 3.3 70B Instruct models.

\begin{table*}
  \centering
  \begin{tabular}{p{4.5cm}p{4.5cm}p{2cm}p{2cm}}
    \toprule
     Feature Identification \& Extraction Model & Fine-tune model  &  ClinStructor finetuned  & Full admission notes  \\
    \midrule
    \multirow{4}{*}{LLaMA 3.3 70B Instruct} & LLaMA-3.1-8B & 0.854  & 0.882 \\
    & Meditron3-8B & 0.860 & 0.889\\
    & LLaMA-3.2-3B & 0.852 & 0.879\\
    & LLaMA-3.2-1B & 0.819 & 0.862 \\
    \midrule
    & Best & 0.860 & 0.889\\ 
    \midrule 
    \multirow{4}{*}{Qwen 2.5 32b Instruct} & Qwen 2.5 7B & 0.869 & 0.882 \\
    & Meditron3-Qwen2.5-7B & 0.860 & 0.881\\
    & Qwen2.5-3B & 0.851 & 0.878\\
    & Qwen2.5-0.5B & 0.815 & 0.856\\
     \midrule
    & Best & 0.869 & 0.882\\ 
    \bottomrule
  \end{tabular}
  \caption{\textbf{Direct Fine-Tuning vs. ClinStructor Fine-Tuning:} As expected, direct fine-tuning performs slightly better than ClinStructor. However, ClinStructor still achieves comparable performance, indicating that it captures most, if not all, of the relevant information.}
  \label{tab:main_experiment}
\end{table*}

\subsection{Stage 2: Feature Extraction}

In this step, our objective is to extract the previously identified features from each patient's admission note. Again, we leverage large language models (LLMs) for this task. For each patient in the dataset, we provide the LLM with the complete admission note along with the set of 50 curated questions obtained from the Feature Identification step. The LLM is instructed to generate an answer for each question based exclusively on the information present in the given admission note. To handle situations where the required information is missing or the question is not applicable, the LLM is explicitly instructed to respond with "N/A".

Consequently, at the end of this step, each patient record—across training, validation, and test sets—is associated with a structured set of 50 question-answer pairs. Importantly, the same set of questions (features) is consistently applied across all examples in the dataset. This representation offers a structured yet flexible format, as the answers retain their original free-text semantic richness while capturing clinically meaningful information in a standardized manner.

\subsection{Stage 3: Fine-tuning} 

In the final stage, we fine-tune a smaller LLM to predict the mortality outcome (alive or deceased). The input to the model is a single concatenated sequence comprising all 50 question-answer pairs for each patient. The model is trained using standard binary cross-entropy loss.

\section{Experiments}\label{sec:experiments}

\subsection{Dataset}

In our experiments, we use the MIMIC-III Clinical Database (version 1.4) and follow the preprocessing steps outlined by \citet{vanAken2021}. Specifically, \citet{vanAken2021} extracted discharge summaries but retained only sections containing information known at the time of admission. This effectively reconstructs admission notes from the original discharge summaries.

After preprocessing, we observe significant class imbalance, with negative cases (patients who survived) considerably outnumbering positive cases (patients who were deceased). While additional negative examples can provide valuable signals during training, they significantly increase computational cost. To achieve a practical balance between computational efficiency and model performance, we randomly subsample the negative examples so that their number matches that of positive cases. This results in a balanced dataset for training, validation, and testing 

We employ AUC-ROC as our primary evaluation metric, chosen for its robustness to class imbalance and its common usage in clinical risk prediction studies. After subsampling, the final dataset comprises 7,078 training examples, 1,038 validation examples, and 2,050 test examples.

\subsection{Direct fine-tuning vs. ClinStructor fine-tuning}

In this section, we compare two methods for training Large Language Models (LLMs) to predict ICU mortality:
\begin{enumerate}
    \item \textbf{Direct fine-tuning}: The model is directly trained on raw, unstructured admission notes.
    \item \textbf{ClinStructor fine-tuning}: The model is trained on structured data, specifically a set of 50 question-answer pairs extracted using our proposed method.
\end{enumerate}

This comparison aims to determine whether structuring clinical text via LLM-driven feature extraction results in any loss of predictive information, if so, how much it affects the downstream prediction performance.

Both fine-tuning approaches use the same optimization setup. We employ Low-Rank Adaptation (LoRA) for parameter-efficient fine-tuning and optimize using binary cross-entropy loss. Models are trained for 5 epochs, and their performance is evaluated every 100 steps using the validation set. For each hyperparameter configuration, we select the checkpoint with the lowest validation loss for final evaluation on the test set. We perform a grid search over two learning rates (1e-4 and 2e-4) and two batch sizes (8 and 16), and report the test results corresponding to the hyperparameter with the best validation performance.

To assess the impact of model size and domain-specific clinical knowledge, we experiment with various LLMs. These include general-purpose models such as Qwen 2.5 (7B, 3B, 0.5B) and LLaMA 3.1 (8B), LLaMA 3.2 (3B, 1B), as well as clinical-specialized variants like Meditron3-Qwen 2.5 (7B) and Meditron3-LLaMA 3.1 (8B). 

We use the LLaMA 70B Instruct and Qwen 2.5 32B Instruct models for the Feature Identification (Stage 1) and Feature Extraction (Stage 2) phases described previously. These larger models are solely used for structuring the data and are not involved in any downstream fine-tuning or prediction tasks, ensuring fairness in our experimental comparison.

\subsection{Results}

Table~\ref{tab:main_experiment} shows that ClinStructor fine-tuning achieves performance competitive with that of Direct fine-tuning on raw admission notes. Specifically, the performance difference in AUC is less than 3\% for the best-performing LLaMA models and less than 2\% for the best-performing Qwen models. Although there is a slight decrease in performance, in ClinStructor, we are able to control exactly what information is used for making the prediction.

\section{Analysis}\label{sec:analysis}



\begin{figure*}[htbp]
    \centering
    \begin{minipage}[t]{0.48\textwidth}
        \centering
        \includegraphics[width=\linewidth]{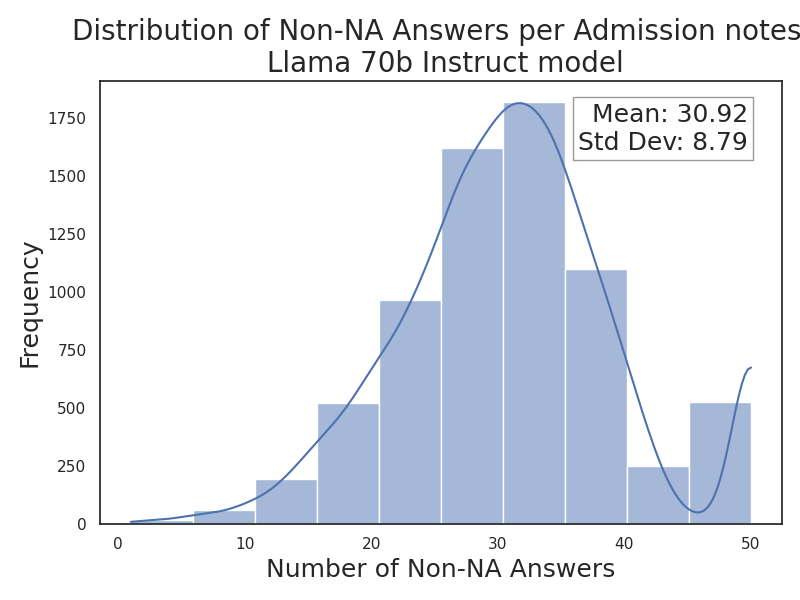}
    \end{minipage}
    \hfill
    \begin{minipage}[t]{0.48\textwidth}
        \centering
        \includegraphics[width=\linewidth]{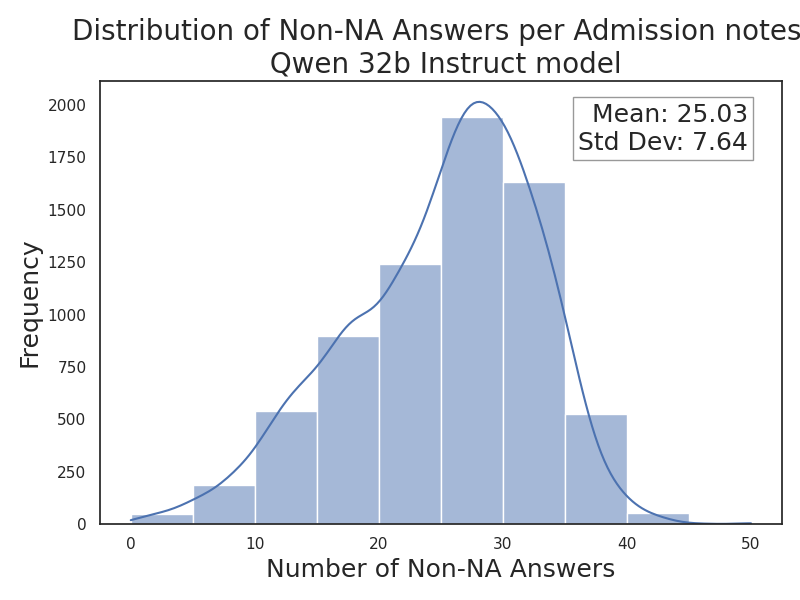}
    \end{minipage}
    \caption{\textbf{Effective Number of Questions:} The plot shows the distribution of the effective number of questions—i.e., the number of questions with non-"N/A" answers, for both Qwen-32B and LLaMA-70B models. Note that approximately 40\% of the answers from LLaMA-70B and 50\% from Qwen-32B are "N/A".}
    \label{fig:num_non_na}
\end{figure*}

\subsection{Effective number of features}

Although 50 question-answer pairs per admission note may appear sufficiently large to capture most relevant clinical information, not every question is applicable to every patient. In practice, many of the answers returned for a given patient are "N/A"—indicating either that the question is not relevant or that the necessary information is not present in the admission note. As shown in Figure~\ref{fig:num_non_na}, for LLaMA 70B and Qwen 32b models, nearly 40\% and 50\% of the answers, respectively, are "N/A", thereby reducing the effective number of useful question-answer pairs per patient.

\subsection{Effect of number of Questions:}

Intuitively, the more number of questions we select, the more information we can retain and hence the lesser the performance loss. However, using more questions increases the difficulty of interpretability and the computational cost. To understand the tradeoff between performance and number of questions, we conduct experiment with various numbers of questions. Instead of selecting top 50 questions, we choose top k questions for K = 10, 20, 30, 40 and 50. We fine-tune Qwen 2.5 7B model and LLaMA 3.1 8B models, for each, we chose the best performing hyperparameter from Table~\ref{tab:main_experiment} and used the same for all different values of K. 

From Table~\ref{tab:num_ft}, we can see that while there is an overall trend (although noisy) of increase in performance with number of questions, even with fewer questions, like 10 for LLaMA and 20 for Qwen, the performance remains close to that of using all the 50 questions.

\begin{table*}
  \centering
  \begin{tabular}{lcccccc}
    \toprule
     Model & Top 10 & Top 20 & Top 30 & Top 40 & Top 50 \\
     \midrule
     Qwen2.5 7B & 0.741 & 0.849 & 0.850 & 0.848 & 0.869 \\
     LLaMA 3.1 8B & 0.849 & 0.847 & 0.853 & 0.856 & 0.854 \\
    \bottomrule
  \end{tabular}
  \caption{\textbf{Number of Features vs. Performance:} AUC-ROC of Qwen2.5-7B and LLaMA 3.1-8B ClinStructor fine-tuned models using only the top-K features. For feature extraction, we use Qwen-32B Instruct and LLaMA3.3-70B Instruct models for Qwen2.5 7B and LLaMA 3.1 8B, respectively.}\label{tab:num_ft}
\end{table*}

\section{Discussion}\label{sec:discussion}

\textbf{Performance.} While ClinStructor provides greater control over which features are used for prediction, this often comes at the cost of reduced performance. One straightforward way to mitigate this trade-off is to use more powerful language models, such as GPT-4.5 or Gemini Pro, during the feature identification and extraction steps. Due to licensing restrictions with the MIMIC dataset, we do not use OpenAI models in our current experiments. However, future work can explore the use of Gemini models through Vertex AI, which complies with MIMIC’s data usage policies. Additionally, further improvements may be achieved by tuning more hyperparameters or significantly increasing the number of questions used in the pipeline (e.g., scaling to hundreds of questions).\\

\noindent\textbf{Interpretability.} ClinStructor marks an important first step toward building interpretable clinical models. In its current form, it enables us to answer the question: Which features were used to make a prediction? However, it does not yet provide insights into how each feature influences the prediction outcome. While techniques like saliency maps or attention-based visualizations offer partial explanations, they are often unreliable or lack fidelity to the model’s actual decision-making process. A more principled extension—drawing inspiration from Neural Additive Models (NAMs)~\citep{agarwal2021neuraladditivemodelsinterpretable}—could improve interpretability. NAMs are a class of models that combine the interpretability of generalized additive models with the flexibility of neural networks by learning a separate sub-network for each input feature. Specifically, instead of regular fine-tuning in Stage 3 of ClinStructor, we could design a NAM-like architecture where each question (feature) is processed independently to produce a logit, and the final prediction is obtained through a linear combination of these logits. This approach would allow us to directly quantify the contribution of each feature to the final prediction. Another possible direction is to guide the LLM to generate questions with strictly numerical or categorical answers, enabling the use of simpler, inherently interpretable models on top of these features. However, this approach may require a significantly larger number of questions, which would increase computational costs and potentially reduce performance, as it limits the model’s ability to leverage pre-trained knowledge (which the current fine-tuned models can leverage). \\

\section{Conclusion}\label{sec:conclusion}

In this work, we introduce ClinStructor, an LLM-based pipeline that converts unstructured clinical notes into structured representations, which are then used to train a predictive model. This approach enhances interpretability and provides control over which features contribute to a prediction—both of which are critical in clinical decision-making.

We evaluate ClinStructor on the MIMIC ICU mortality prediction task using admission notes. Our proposed method results in only a modest drop in AUC (2–3\%) while offering greater control over the features used in prediction. Further analysis reveals that, on average, only 50–60\% of the generated questions are relevant for a given patient's notes. Interestingly, using just the top 10 most informative questions yields performance close to that of using all 50.

ClinStructor also addresses common challenges in clinical machine learning. By manually reviewing and selecting the final set of questions, we can reduce the risk of unintended biases. The consistent input format—50 standardized questions and answers—improves generalizability across different EHR systems and further enhances interpretability. Overall, this work demonstrates the promise of using large language models not only as predictive tools but also as enablers of transparent, robust, and controllable clinical machine learning systems.

\section{Limitations}  
Our proposed approach results in a slight decrease in performance. While the outputs are interpretable to some extent, full transparency is lacking—we can identify which features are used in making a prediction, but not how they influence the outcome. The approach relies on large language models (LLMs) to transform data, which introduces concerns about potential hallucinations. Moreover, leveraging more powerful LLMs may require sending data to external services, raising additional challenges related to privacy and control. Fine-tuning these models locally also demands substantial computational resources.

\section*{Ethical considerations}
This work involves healthcare-related tasks, a domain that requires careful handling due to its sensitive nature. To the best of our knowledge, all experiments were conducted using the latest PII-deidentified data from the MIMIC database. We strictly followed the guidelines provided by MIMIC and ensured that no OpenAI models were used. All large language models (LLMs) were hosted locally to minimize the risk of data leakage and to maintain data privacy.

The MIMIC dataset is approved for research use, and all the LLMs employed in this study are also permitted for research purposes.




\bibliography{custom}

\newpage
\appendix

\section{50 Questions from Qwen 32b model}

\begin{lstlisting}[
  basicstyle=\ttfamily,
  breaklines=true,
  breakatwhitespace=true,
  keepspaces=true
]
1. what is the patient’s age?
2. what medications was the patient on at the time of admission?
3. what is the patient’s current vital signs?
4. what is the patient’s physical exam status?
5. what are the patient’s significant past medical conditions?
6. what is the patient’s history of previous surgeries?
7. what is the patient’s history of recent infections?
8. what is the patient’s family history?
9. what is the patient’s history of allergies?
10. what is the patient’s history of recent lab results?
11. what is the patient’s history of social habits?
12. what is the patient’s history of renal disease?
13. what is the patient’s primary diagnosis?
14. what is the patient’s history of recent changes in respiratory status?
15. what is the patient’s recent imaging results?
16. what is the patient’s current mental status?
17. what is the patient’s history of cardiovascular events?
18. what is the patient's recent nutritional status?
19. what is the patient’s history of hospitalizations?
20. what is the patient’s current cardiovascular status?
21. what is the patient’s current neurological status?
22. what is the patient’s history of chronic conditions?
23. what is the patient’s history of respiratory conditions?
24. what is the patient’s current pain level?
25. what is the patient’s history of recent changes in liver function?
26. what is the patient’s history of neurological disorders?
27. what is the patient’s current oxygen saturation?
28. what is the patient’s history of trauma?
29. what is the patient’s current level of consciousness?
30. what is the patient’s blood pressure?
31. what is the patient’s history of mental health conditions?
32. what is the patient’s heart rate?
33. what is the patient’s history of metabolic diseases?
34. what is the patient’s history of chronic diseases?
35. what is the patient’s chief complaint?
36. what is the patient’s respiratory rate?
37. what is the patient’s history of gastrointestinal issues?
38. what is the patient’s history of recent changes in fluid balance?
39. what is the patient’s alcohol consumption?
40. what is the patient’s current mobility status?
41. what is the patient’s history of smoking?
42. what is the patient’s history of cancer?
43. what is the patient's recent functional status?
44. what is the patient’s current treatment plan?
45. what is the patient’s history of hypertension?
46. what is the patient’s history of infectious diseases?
47. what is the patient’s history of substance use?
48. what is the patient’s recent activity level?
49. what is the patient’s history of recent treatments?
50. what is the patient’s history of previous hospital admissions?
\end{lstlisting}

\section{Question Generation Prompts}
\label{app:gen_prompts}

\begin{lstlisting}[
    basicstyle=\ttfamily,        % Monospaced, NO size change
    breaklines=true,           % Wraps long lines
    breakatwhitespace=true,    % Wraps at spaces
    keepspaces=true,           % Preserves your indentation
    alsoletter={_,\{,\}},
    literate={’}{'}1  % Fixes: patient’s
             {“}{"}1  % Fixes: “What
             {”}{"}1  % Fixes: age?”
]
system_prompt

You are an expert feature engineer with specialization in clinical and medical domains. You are helping to design features for a mortality prediction model.

--------------------
Instruction:

Your task is to define useful features for predicting in-hospital mortality from ICU admission notes.

Do the following:

1. Write 20 generalizable and clinically meaningful questions that could be answered from admission notes. Answer to these questions should be good predictors of mortality.
2. Assign a short feature name (keyword) to each question.
3. Rate each feature’s importance from 0 (not useful) to 1 (highly predictive).

Guidelines:

1.  Avoid yes/no questions. Use open-form questions (e.g., prefer “What is the patient’s age?” over “Is the patient older than 65?”).

Provide your answer in the given JSON output format. 

Example of one question_info
  "question": "What is the patient’s age?",
  "keyword": "patient_age",
  "importance": 0.8


Patient Admission Notes: 
{patient_notes}


--------------------
Constraint Output Schema

json_schema = {
  "name": "questions_generation",
  "description": "20 questions that serves as a feature for mortality prediction.",
  "schema": {
    "type": "object",
    "properties": {
      "question_info": {
        "type": "array",
        "items": {
          "type": "object",
          "properties": {
            "question": {
              "type":"string",
              "description": "A clinical question that can be answered from the note"
            },
            "keyword": {
              "type": "string",
              "description": "A concise name for the feature"
            },
            "importance": {
              "type": "number",
              "description": "Feature importance score from 0 (low) to 1 (high)"
            }
          },
          "required": ["question", "keyword", "importance"]
        }
      }
    },
    "required": ["question_info"]
  }
}
\end{lstlisting}

\section{Answer Generation Prompts}

\begin{lstlisting}[
    basicstyle=\ttfamily,        % Monospaced, NO size change
    breaklines=true,           % Wraps long lines
    breakatwhitespace=true,    % Wraps at spaces
    keepspaces=true,           % Preserves your indentation
    alsoletter={_,\{,\}},
    literate={’}{'}1  % Fixes: patient’s
             {“}{"}1  % Fixes: “What
             {”}{"}1  % Fixes: age?”
]
system_prompt_answergen: 

You are an expert in the clinical and medical domain, with expertise in analyzing and answering any questions based on clinical notes.

-----------------------


json_schema_answergen = {
    "name": "answer_50",
    "description": "Answer to the given 50 questions.",
    "schema": {
                  "type": "object",
                  "properties": {
                    "Q1": { "type": "string" },
                    "Q2": { "type": "string" },
                    "Q3": { "type": "string" },
                    .......
                    .......
                    "Q49": { "type": "string" },
                    "Q50": { "type": "string" }
                  },
                  "required": [
                    "Q1", "Q2", "Q3", .......... "Q49", "Q50"
                  ]
                }
}


\end{lstlisting}

\section{More details on Clustering}

When generating the 50 questions, the LLM outputs not only the question text but also a corresponding feature name and importance score. For example:
\begin{lstlisting}[
    basicstyle=\ttfamily,
    keepspaces=true,
    breaklines=true,           
    alsoletter={_,\{,\}}
]
{'question': 'What is the patient's age?',
 'keyword': 'patient_age',
 'importance': 0.8}
\end{lstlisting}

We observed that while the phrasing of the questions may vary, the feature names are often quite consistent. For instance, all of the following questions map to the same feature name \texttt{past\_medical\_conditions} and therefore belong to the same cluster:
\begin{itemize}
    \item what are the patient's past medical conditions?
    \item what are the patient's significant past medical conditions?
    \item what are the patient's past medical conditions?
    \item what are the patient's significant past medical conditions?
    \item what are the significant past medical conditions?
\end{itemize}
We further improve clustering by working in the reverse direction as well. For example, for the question \textit{what is the patient's laboratory test results?} the LLM sometimes produced different feature names such as \texttt{ab\_results}, \texttt{lab\_test\_results}, and \texttt{lab\_tests}. In such cases, we group all of these feature names and their associated questions into the same cluster. This clustering procedure can be formalized as finding the connected components of a graph, where nodes are feature names and questions, and edges connect each feature name to its corresponding questions. Implementation is straightforward using NetworkX, as shown below:

\begin{lstlisting}[
    language=Python,
    basicstyle=\ttfamily,       
    commentstyle=\ttfamily, 
    breaklines=true,
    breakatwhitespace=true,
    keepspaces=true,
    alsoletter={_} % Treats underscore as a normal letter
]
import networkx as nx

def build_keyword_question_clusters(
keyword2question, question2keyword):
    G = nx.Graph()

    # Add edges from both dictionaries
    for kw, questions in keyword2question.items():
        for q in questions:
            G.add_edge(kw, q)

    for q, keywords in question2keyword.items():
        for kw in keywords:
            G.add_edge(q, kw)

    # Extract connected components
    components = list(nx.connected_components(G))

    # Optional: extract just keyword clusters
    keyword_clusters = [
        [node for node in comp if node in keyword2question]
        for comp in components
    ]

    return keyword_clusters, components
\end{lstlisting}
Finally, for each connected component, we select the most frequent question to represent the cluster.

\end{document}